\pdfoutput=1
\documentclass[12pt, a4paper]{article}

\usepackage[utf8]{inputenc}
\usepackage[T1]{fontenc}
\usepackage{lmodern}

\usepackage[top=2.5cm, bottom=2.5cm, left=2.5cm, right=2.5cm]{geometry}

\usepackage{amsmath}
\usepackage{amssymb}
\usepackage{bm}

\usepackage{graphicx}
\usepackage{subcaption}
\usepackage{booktabs}
\usepackage{multirow}
\usepackage{array}
\usepackage{float}

\usepackage[colorlinks=true, linkcolor=blue, citecolor=blue, urlcolor=blue]{hyperref}

\usepackage{parskip}
\usepackage[numbers, sort&compress]{natbib}

\usepackage[ruled, linesnumbered]{algorithm2e}

\usepackage{xcolor}
\definecolor{notecolour}{rgb}{0.6,0.0,0.0}

\newcommand{\figplaceholder}[2][0.85\textwidth]{%
  \fbox{\begin{minipage}[c][5cm][c]{#1}%
    \centering\textcolor{gray}{\textit{[Figure pending: #2]}}%
  \end{minipage}}%
}
\newcommand{\safeinclude}[3][width=0.85\textwidth]{%
  \IfFileExists{#2}{\includegraphics[#1]{#2}}{\figplaceholder{#3}}%
}

\newcommand{\zk}{\mathbf{z}(k)}
\newcommand{\zkp}{\mathbf{z}(k{+}1)}
\newcommand{\uk}{\mathbf{u}(k)}
\newcommand{\Lewc}{\mathcal{L}_{\mathrm{EWC}}}
\newcommand{\Lphys}{\mathcal{L}_{\mathrm{phys}}}
\newcommand{\Ldata}{\mathcal{L}_{\mathrm{data}}}

\begin{document}

\thispagestyle{empty}
\vspace*{2cm}
\begin{center}

{\Large \textbf{World Model for Battery Degradation Prediction\\[4pt]
Under Non-Stationary Aging}}

\vspace{1.5cm}

{\large Kai Chin Lim\textsuperscript{1}, Khay Wai See\textsuperscript{1}}

\vspace{0.5cm}

{\normalsize \textsuperscript{1}Independent Researcher, Australia}

{\small \texttt{kc23.lim@gmail.com, khaywai29@gmail.com}}

\vspace{1.5cm}

\textbf{Abstract}

\vspace{0.5cm}
\end{center}

\begin{quote}
Degradation prognosis for lithium-ion cells requires forecasting the
state-of-health (SOH) trajectory over future cycles. Existing
data-driven approaches can produce trajectory outputs through direct
regression, but lack a mechanism to propagate degradation dynamics
forward in time. This paper formulates battery degradation prognosis as
a world model problem, encoding raw voltage, current, and temperature
time-series from each cycle into a latent state and propagating it
forward via a learned dynamics transition to produce a future trajectory
spanning 80 cycles. To investigate whether physics constraints
improve the learned dynamics, a monotonicity penalty derived from
irreversible degradation is incorporated into the training loss. Three
configurations are evaluated on the Severson
LiFePO\textsubscript{4} (LFP) dataset of 138~cells. Iterative rollout
halves the trajectory forecast error compared to direct regression from
the same encoder. The physics constraint improves prediction at the
degradation knee without changing aggregate accuracy.

\end{quote}

\vspace{1cm}
\begin{center}
\end{center}

\newpage


\section{Introduction}
\label{sec:introduction}

Battery degradation prognosis requires forecasting the
state-of-health (SOH) trajectory over future cycles from available
cycling measurements. Existing data-driven approaches such as Long Short-Term Memory (LSTM),
Gated Recurrent Unit (GRU), and transformer models can map a window of cycle measurements to
SOH estimates and can produce trajectory outputs through direct
regression from a learned representation. These approaches lack a
mechanism to propagate degradation dynamics forward in time, and the
resulting trajectory predictions learn an average slope applied
uniformly across all horizons.

This paper formulates battery degradation prognosis as a world model
problem. The architecture encodes raw voltage, current, and temperature
time-series from each cycle into a latent state via a 1-D convolutional
encoder and a PatchTST transformer, then propagates the latent state
forward through a learned dynamics transition. A shared decoder maps
each propagated state to an SOH value, producing both a current estimate
and a future trajectory spanning 80 cycles from a single architecture
trained end to end. To investigate whether physics constraints
improve the learned dynamics, a monotonicity penalty derived from
irreversible degradation is incorporated into the training loss.

Three world model configurations are evaluated on the Severson LFP
dataset~\cite{severson2019} of 138~cells aged under fast charging
protocols: rollout with physics constraint, rollout without physics
constraint, and the encoder backbone evaluated with direct regression. Two control
tests complement the study: a plain LSTM baseline and a continual
learning configuration using Elastic Weight Consolidation (EWC) across
manufacturing batches.

The ablation identifies iterative latent rollout as the essential
component for trajectory prognosis. Rollout halves the forecast error
at horizon~5 compared to direct regression from the same encoder, and
the error grows with horizon as expected from genuine iterative
propagation. Direct regression produces flat error across all horizons,
consistent with a learned average slope. The physics constraint improves
prediction at the degradation knee without changing aggregate accuracy. Continual learning with EWC across manufacturing
batches yields 3.3 times worse accuracy than joint training,
confirming that EWC provides no benefit when sequential batches share
the same data distribution. Fisher information computed after
convergence yields near zero values, making EWC functionally inactive.
Computing Fisher at a fixed midtraining epoch resolves this.

Section~\ref{sec:related} reviews related work.
Sections~\ref{sec:method} and~\ref{sec:setup} describe the
architecture and experimental setup.
Section~\ref{sec:results} presents the results.
Section~\ref{sec:discussion} discusses the findings.

\section{Related Work}
\label{sec:related}

\subsection{Data-Driven SOH Estimation}

Recurrent neural networks, in particular LSTM and GRU, have been widely
applied to battery SOH estimation from cycling data. These models accept
a window of cycle measurements and output a scalar SOH estimate.
Performance on held out cells from the same dataset is competitive with
model-based methods. These architectures assume that training and test
data are drawn from the same distribution, an assumption that breaks
down in sequential aging scenarios.

Severson et al.~\cite{severson2019} demonstrated that early-cycle
discharge features contain sufficient information to predict end-of-life
for LFP cells, establishing the Severson dataset as a standard benchmark
for battery lifetime prediction. Their work addressed a different task
from per-cycle SOH tracking, predicting total cycle life from the first
100~cycles. The present work uses the same dataset but targets per-cycle
SOH estimation and future trajectory forecasting across the full aging
lifetime. Gu et al.~\cite{gu2022cnntransformer} proposed a CNN-Transformer
for SOH estimation, combining convolutional local feature extraction
with transformer global attention.
Yunusoglu et al.~\cite{yunusoglu2025llm} apply a transformer
to battery capacity estimation, achieving competitive accuracy on
single step SOH outputs.

Model-based approaches including Kalman filter variants have been
applied to battery state estimation~\cite{lim2016fading}. These require
accurate system models and manual tuning for each cell chemistry.
Data-driven methods learn directly from cycling measurements.

\subsection{Physics-Informed Machine Learning for Batteries}

Physics-informed neural networks (PINNs)~\cite{raissi2019} incorporate
physical constraints into the training loss as soft penalties, without
requiring the network to explicitly solve the governing equations at
inference time. Applied to batteries, constraints derived from
electrochemical models can regularise learned trajectories toward
physically consistent behaviour. Wang et al.~\cite{wang2024pinn}
demonstrated that embedding electrochemical degradation dynamics
directly into the loss function stabilises SOH estimation and prognosis
from partial charge/discharge curves. Borah et al.~\cite{borah2024synergy}
survey the integration of physics-based and machine learning models
across state estimation, health monitoring, and safety diagnostics,
identifying resistance growth modelling as a key open area.

The Single Particle Model (SPM)~\cite{doyle1993} reduces the
electrochemical dynamics of a lithium-ion cell to a single spherical
particle per electrode. Under the SPM, internal resistance grows with
cycling according to a power-law relationship with SOH:
$R/R_0 \sim (1/\mathrm{SOH})^\gamma$, where $R_0$ is the initial
resistance and $\gamma$ is a chemistry-specific exponent. For LFP
chemistry, $\gamma \approx 0.75$~\cite{prada2012}. This relationship
provides a direct constraint between observable resistance and predicted
SOH.

\subsection{Continual Learning}

Catastrophic forgetting~\cite{mccloskey1989} describes the tendency of
neural networks to overwrite weights learned on one task when trained
sequentially on a second task. Elastic Weight Consolidation
(EWC)~\cite{ewc2017} addresses forgetting by adding a quadratic penalty
that anchors weights important to prior tasks. Importance is estimated
via the diagonal of the Fisher information matrix, approximated as mean
squared gradients over the prior task data. EWC has been applied in
computer vision and natural language processing.
Sayed et al.~\cite{sayed2025cl} survey continual learning methods for
energy management systems, covering replay, regularisation, and
architectural approaches. Application of EWC to sequential battery
aging stages has not appeared in the literature.

\subsection{World Models}

The world model formulation~\cite{sutton1991dyna, ha2018world} posits an
internal model of the environment that predicts future states from
current state and action: $s_{t+1} = f(s_t, a_t)$. This framing has
been applied primarily in model-based reinforcement learning for
planning and imagination. Mao et al.~\cite{mao2024worldmodel} extend
world models to physical systems, aligning latent representations with
physical quantities and constraining their evolution through known
physics equations in robotic control. The formulation does not prescribe a specific encoder or transition
architecture; Ha and Schmidhuber~\cite{ha2018world} use a VAE with an
RNN, while Mao et al.~\cite{mao2024worldmodel} use a VQ-VAE with
structured dynamics. World models have not been
applied to battery degradation prognosis.

\section{Methodology}
\label{sec:method}

\subsection{Problem Formulation}

Let a battery cell produce a sequence of per-cycle measurements. At
cycle~$k$, the raw observation consists of three time-series recorded
during discharge: voltage $V(t)$, current $I(t)$, and temperature
$T(t)$, each sampled at ${\sim}1000$ timesteps. Discharge capacity and internal resistance require lab grade equipment
and are excluded from the model input.

The model receives a window of $W = 30$ consecutive cycles and
produces two outputs:
\begin{align}
  \hat{s}(k) &= \text{current SOH at cycle } k
    \quad (\text{scalar}), \\
  \hat{\mathbf{s}}(k{+}1:k{+}H) &= \text{predicted SOH for the next }
    H = 80 \text{ cycles} \quad (\text{vector} \in \mathbb{R}^H).
\end{align}

SOH is defined as the ratio of current discharge capacity to the
reference capacity at cycle~2, following the normalisation convention
of~\cite{severson2019}.

\subsection{World Model Architecture}

\begin{figure}[htbp]
  \centering
  \safeinclude[width=0.92\textwidth]{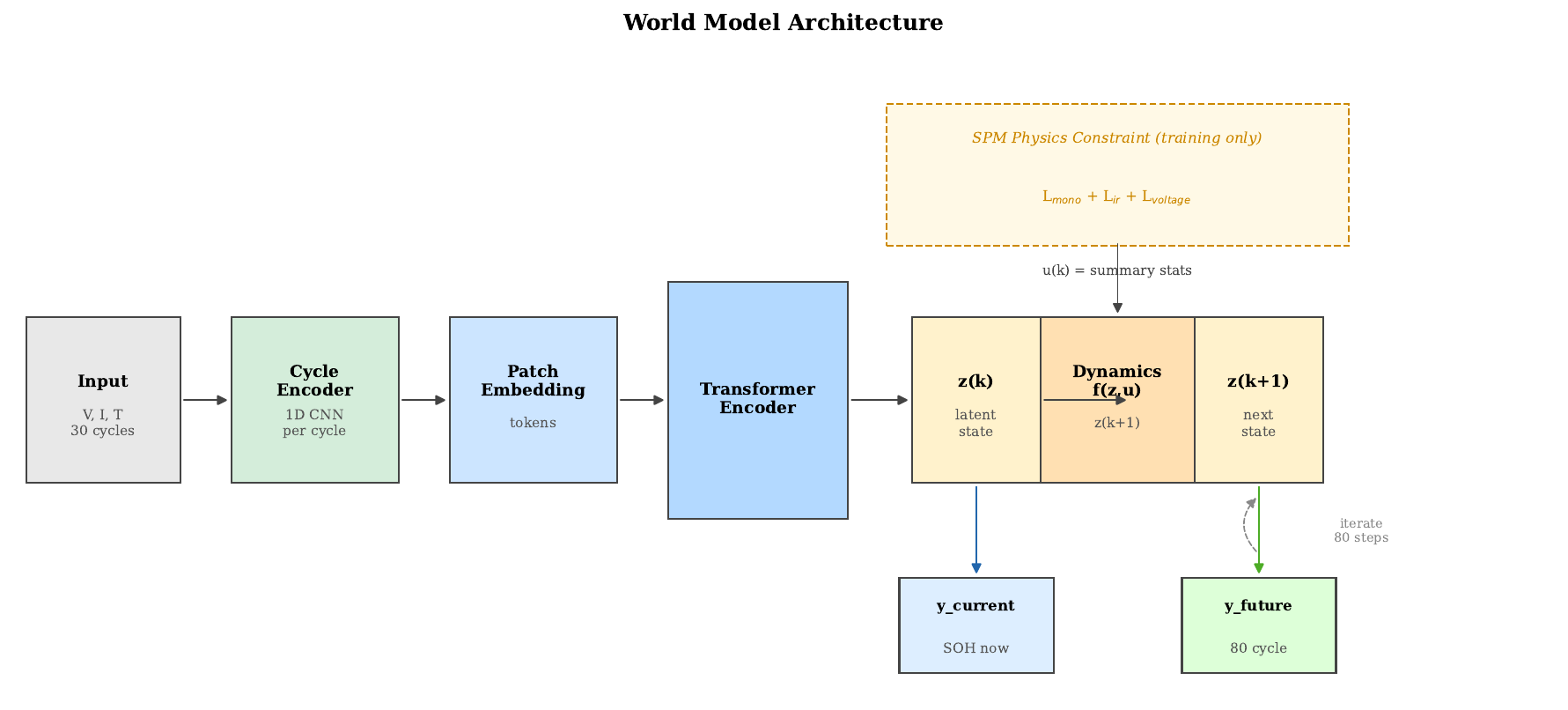}{Architecture diagram}
  \caption{World model architecture. Raw V/I/T time-series per cycle
    are encoded by a shared 1-D CNN, then processed by PatchTST to
    produce latent state $\zk$. The dynamics module rolls out future
    latent states and a shared head decodes SOH at each step. The physics
    constraint operates during training only.}
  \label{fig:architecture}
\end{figure}

The architecture is illustrated in Figure~\ref{fig:architecture}. Four
components transform the input window into the dual output.

\textbf{Cycle Encoder.} Each cycle's raw time-series
$[V(t), I(t), T(t)] \in \mathbb{R}^{3 \times T_{\max}}$ (where
$T_{\max} = 1000$) is processed independently by a shared 1-D
convolutional neural network. The encoder consists of three Conv1d
layers (channels 32$\to$64$\to$128, kernels 7/5/3, stride 2) with
batch normalisation and ReLU activation, followed by adaptive average
pooling and a linear projection to a $d$-dimensional embedding
$\mathbf{e}(k) \in \mathbb{R}^d$, where $d = 64$. This produces a
fixed-size representation per cycle regardless of the raw time-series
length, capturing discharge curve shape, voltage plateau
characteristics, and thermal signatures.

\textbf{PatchTST Encoder.} The sequence of cycle embeddings
$[\mathbf{e}(k{-}W{+}1), \ldots, \mathbf{e}(k)] \in
\mathbb{R}^{W \times d}$ is divided into patches of length $P = 6$
cycles with stride $S = 3$, following the PatchTST
formulation~\cite{patchtst2023}. Each patch is linearly projected to a
$d$-dimensional token. Sinusoidal positional encodings are added. A
multihead self-attention transformer encoder with $L = 3$ layers,
$N_h = 4$ heads, and feedforward dimension $d_{\mathrm{ff}} = 256$
processes the token sequence, learning relationships between patches
at different temporal positions. The output is pooled via adaptive
average pooling to a single vector $\zk \in \mathbb{R}^d$,
representing the latent degradation state at cycle~$k$.

\textbf{Dynamics Transition.} The action vector $\uk$ at the last
observed cycle encodes the charging current ($I_{\mathrm{mean}}$),
the operating condition that varies across cells in the fast charging
dataset. A two-layer MLP with residual connection
computes the next latent state:
\begin{equation}
  \zkp = \zk + \mathrm{MLP}\bigl([\zk \,\|\, \uk]\bigr),
  \label{eq:dynamics}
\end{equation}
where $[\cdot \| \cdot]$ denotes concatenation. The residual
connection enforces that $\zkp$ departs from $\zk$ by a learned
increment, reflecting the physical intuition that degradation is a
continuous process with bounded per-cycle change. Iterating
Equation~\eqref{eq:dynamics} for $H$~steps produces the rollout
sequence $\{\mathbf{z}(k{+}1), \ldots, \mathbf{z}(k{+}H)\}$.

\textbf{Output Head.} A shared decoder maps latent states to SOH:
\begin{align}
  \hat{s}(k) &= \mathrm{head}_{\mathrm{current}}(\zk), \\
  \hat{s}(k{+}h) &=
    \mathrm{head}_{\mathrm{current}}(\mathbf{z}(k{+}h)),
    \quad h = 1, \ldots, H.
\end{align}
A single shared head (two-layer MLP with ReLU) maps any latent
vector to a scalar SOH estimate. The future trajectory is obtained
by applying this head to each step of the dynamics rollout.

\subsection{Training Objective}

The total training loss combines up to three terms:
\begin{equation}
  \mathcal{L} = \Ldata + \lambda_{\mathrm{phys}} \Lphys +
                \lambda_{\mathrm{EWC}} \Lewc,
  \label{eq:loss}
\end{equation}
where $\lambda_{\mathrm{phys}} = 0.1$ and $\lambda_{\mathrm{EWC}} =
0.4$. Each term can be independently disabled for ablation
as described in Section~\ref{sec:results}.

\textbf{Data loss.} Mean squared error on current SOH and future
trajectory:
\begin{equation}
  \Ldata = \mathrm{MSE}\bigl(\hat{s}(k),\, s(k)\bigr) +
           \mathrm{MSE}\bigl(\hat{\mathbf{s}}(k{+}1:k{+}H),\,
           \mathbf{s}(k{+}1:k{+}H)\bigr).
\end{equation}

\subsection{Physics Constraint}

The physics constraint $\Lphys$ enforces physically consistent
trajectory predictions through three terms:

\textbf{Monotonicity loss} $\mathcal{L}_{\mathrm{mono}}$.
SOH degradation is irreversible under normal cycling. The predicted
future trajectory must be nonincreasing:
\begin{equation}
  \mathcal{L}_{\mathrm{mono}} =
    \frac{1}{H-1} \sum_{h=1}^{H-1}
    \max\!\bigl(0,\; \hat{s}(k{+}h{+}1) - \hat{s}(k{+}h)
    + \varepsilon\bigr)^2,
\end{equation}
where $\varepsilon = 0.005$ is a tolerance for measurement noise.

\textbf{Resistance to SOH consistency loss} $\mathcal{L}_{\mathrm{ir}}$.
The SPM predicts that internal resistance grows with cycling as:
\begin{equation}
  \frac{R}{R_0} \sim \left(\frac{1}{\mathrm{SOH}}\right)^\gamma,
  \quad \gamma = 0.75 \text{ (LFP, A123 cell)},
  \label{eq:spm}
\end{equation}
where $R_0$ is the initial resistance. The SOH implied by
the observed resistance ratio is:
\begin{equation}
  s_{\mathrm{IR}} = \left(\frac{R_0}{R_{\mathrm{last}}}\right)^{1/\gamma}.
  \label{eq:soh_ir}
\end{equation}

The resistance to SOH consistency loss penalises the deviation of the
predicted current SOH from the SPM-implied value:
\begin{equation}
  \mathcal{L}_{\mathrm{ir}} = \mathrm{MSE}\bigl(\hat{s}(k),\,
    s_{\mathrm{IR}}\bigr).
\end{equation}

\textbf{Voltage consistency loss} $\mathcal{L}_{\mathrm{voltage}}$.
A relative check on terminal voltage consistency with observed current
and resistance, serving as a structural regulariser.

The composite physics loss is:
\begin{equation}
  \Lphys = \mathcal{L}_{\mathrm{mono}} + \mathcal{L}_{\mathrm{ir}}
           + \mathcal{L}_{\mathrm{voltage}}.
\end{equation}

The effect of enabling and disabling $\Lphys$ is reported in
Section~\ref{sec:results}.

\subsection{Elastic Weight Consolidation}

EWC~\cite{ewc2017} prevents catastrophic forgetting by penalising
changes to weights that were important for prior data. After training
on data subset~$t$, the importance of each weight $\theta_i$ is
estimated by the diagonal Fisher information:
\begin{equation}
  F_i \approx \frac{1}{N} \sum_{n=1}^{N}
    \left(\frac{\partial \Ldata^{(n)}}{\partial \theta_i}\right)^2,
  \label{eq:fisher}
\end{equation}
where the sum is over $N$ mini-batches. During training on the next
subset, the EWC penalty is:
\begin{equation}
  \Lewc = \sum_i F_i \left(\theta_i - \theta_i^*\right)^2,
  \label{eq:ewc}
\end{equation}
where $\theta_i^*$ are the weights at the end of the previous subset.

\textbf{Fisher timing.} Fisher must be computed before the model
reaches convergence. At
convergence, gradients approach zero and $F_i \approx 0$ for all~$i$,
making $\Lewc$ functionally inactive. The observed ratio
$\Lewc / \Ldata \approx 10^{-4}$ when Fisher is computed
postconvergence confirms that the penalty has no effect. Fisher is computed at epoch~10 during each training
phase, when gradients remain nontrivial. This is nonstandard
relative to the original EWC formulation but is necessary for the
mechanism to activate.

\textbf{Deployment framing.} In the primary training configuration
described in Section~\ref{sec:setup}, all data is available simultaneously and
EWC is not used. EWC is evaluated separately in a batch-staged
configuration that simulates deployment where battery populations
arrive sequentially and historical data is discarded after each batch.

\subsection{Model Variants and Controls}

Three configurations form the architecture ablation, isolating the
contribution of the dynamics rollout and the physics constraint:

\begin{itemize}
  \item \textbf{PIWM} (Physics-Informed World Model):
    $\Ldata + \lambda_{\mathrm{phys}} \Lphys$. Dynamics rollout active.
    Physics active. Single pass joint training.

  \item \textbf{WM} (World Model): $\Ldata$ only. Dynamics rollout
    active. Physics disabled. Single pass joint training. Tests whether
    the physics constraint contributes beyond the data-driven objective.

  \item \textbf{CNN-PatchTST}: $\Ldata$ only. Dynamics module removed;
    future trajectory predicted by direct MLP regression from~$\zk$ in
    a single forward pass. Single pass joint training. This is the
    shared encoder backbone (1-D CNN cycle encoder + PatchTST
    transformer) evaluated without any world model machinery. It
    isolates the contribution of latent space rollout.
\end{itemize}

Two control tests are evaluated separately from the architecture
ablation:

\begin{itemize}
  \item \textbf{PIWM + EWC}: $\Ldata + \lambda_{\mathrm{phys}}
    \Lphys + \lambda_{\mathrm{EWC}} \Lewc$. Same PIWM architecture,
    trained sequentially across three manufacturing batches (batch~1
    $\to$ batch~2 $\to$ batch~3) with EWC protecting previous-batch
    knowledge. Tests whether continual learning improves over joint
    training when sequential batches share the same data distribution.

  \item \textbf{LSTM}: A two-layer LSTM baseline trained on the same
    data provides a comparison against standard recurrent approaches.
    The LSTM produces a single scalar SOH estimate and cannot generate
    trajectory forecasts.
\end{itemize}

\section{Experimental Setup}
\label{sec:setup}

\subsection{Dataset}

Experiments use the Severson et al.\ LFP dataset~\cite{severson2019},
comprising 138 LiFePO\textsubscript{4} cells, specifically A123
APR18650M1A with 1.1~Ah nominal capacity, from three manufacturing
batches. Cells were cycled to end of life under various fast charging
protocols at constant discharge and 30$^\circ$C chamber temperature.
Each cell provides per-cycle raw time-series of voltage, current, and
temperature sampled at approximately 1000 timesteps per cycle, as well
as lab grade measurements of discharge capacity and internal
resistance.

Following~\cite{severson2019}, cycle~1 is excluded due to known data
quality issues. SOH for each cell is normalised by the discharge
capacity at cycle~2, which serves as the reference cycle. Two cells
from batch~1, cells 0 and 18, are excluded due to equipment failure,
leaving 138 usable cells across three batches: 44 from batch~1, 48
from batch~2, and 46 from batch~3.

The 138 cells are split at the cell level into 110 training,
14 validation, and 14 test cells. The split is stratified by minimum
SOH reached, ensuring that cells with deep degradation are
represented in all partitions. No cycle from a test cell appears in
training.

\subsection{Input Representation}

The model input per cycle consists of raw time-series for voltage,
current, and temperature, each padded or truncated to $T_{\max} =
1000$ timesteps. Discharge capacity $Q_d$ and internal
resistance IR, which require lab grade coulomb counting and
impedance measurement respectively, are excluded from the model input.

The charging current $I_{\mathrm{mean}}$ is computed per cycle and
used as the action vector~$\uk$ for the dynamics transition.

Sliding windows of $W = 30$ consecutive input cycles with future
target horizon $H = 80$ cycles are constructed within each cell. No
cross-cell windows are created. The horizon is selected from a sweep
over $H \in \{50, 80, 100\}$: $H{=}80$ yields the lowest overall MAE
(0.006 vs 0.007 for both alternatives) by balancing exposure to the
degradation knee (10.9\% of windows cross SOH~0.95) against rollout
error accumulation at longer horizons.

\subsection{Training Protocol}

The primary training configuration is single pass: all training data
is loaded into a single DataLoader, shuffled every epoch, and
processed with standard gradient descent. The Adam
optimiser~\cite{adam2014} is used with learning rate $10^{-3}$ and
weight decay $10^{-4}$. Early stopping terminates training after
15~epochs without validation MAE improvement, with a maximum of
100~epochs. Gradients are clipped to $\ell_2$ norm~1.0.

\textbf{Imbalance handling.} The SOH distribution is heavily skewed:
83\% of samples fall above SOH~0.95, 11\% between 0.90 and 0.95,
and 6\% between 0.85 and 0.90. Inverse-frequency sampling across four aging stages ensures balanced
representation per batch, preventing the model from learning a
constant healthy-cell predictor.

\textbf{Batch-staged EWC configuration.} For the continual learning
experiment in Section~\ref{sec:results}, the same architecture is
trained sequentially across three manufacturing batches (batch~1
$\to$ batch~2 $\to$ batch~3), with EWC applied between batches.
Fisher information is computed at epoch~10 of each batch phase.
Inverse-frequency sampling is applied within each batch to handle SOH
imbalance. This configuration simulates deployment where each
battery population arrives over time and historical data from
previous batches is discarded.

Hyperparameters are fixed across all variants:
$\lambda_{\mathrm{phys}} = 0.1$, $\lambda_{\mathrm{EWC}} = 0.4$,
window $W = 30$, horizon $H = 80$, batch size 32, $d = 64$.

\subsection{Baselines}

The LSTM baseline is a two-layer LSTM with hidden size 64 and a
two-layer MLP head, trained on the full training set using single
pass training with the same data split. The input dimension matches
the world model's summary features. The LSTM produces a single scalar
SOH estimate per input window and cannot produce multistep future
trajectory predictions. This is an architectural constraint, not a
performance limitation. LSTM results are reported as a control test
in Section~\ref{sec:results}.

\subsection{Evaluation Metrics}

MAE, RMSE, and MAPE are computed per aging stage and overall for all
variants. Aging stages are defined in Table~\ref{tab:stages}. Stage~4 contains
insufficient test samples and is excluded from per-stage reporting.
Stage~2 corresponds to the degradation knee, where capacity fade
accelerates as resistance growth transitions from linear to nonlinear
scaling. The knee region is shaded in Figure~\ref{fig:trajectory}.
Subsequent sections refer to this region as the knee.

\begin{table}[htbp]
\centering
\caption{Aging stage definitions.}
\label{tab:stages}
\footnotesize
\begin{tabular}{lc}
\toprule
Stage & SOH range \\
\midrule
Stage~1 & 1.00--0.95 \\
Stage~2 & 0.95--0.90 \\
Stage~3 & 0.90--0.85 \\
Stage~4 & below 0.85 \\
\bottomrule
\end{tabular}
\end{table}

Future trajectory accuracy is evaluated at horizons $h \in
\{5, 10, 20, 50\}$ cycles ahead for the world model variants.

All evaluation uses the 14 held out test cells. The test set
is not accessed during training or hyperparameter selection.

\section{Results}
\label{sec:results}

\subsection{Architecture Ablation}

Table~\ref{tab:ablation} reports MAE, RMSE, and MAPE on the 14 test
cells, broken down by aging stage, for the three architecture variants.

\begin{table}[htbp]
\centering
\caption{Architecture ablation on the test set. All variants use
  single pass joint training. Lowest values per column are in bold.}
\label{tab:ablation}
\resizebox{\columnwidth}{!}{%
\footnotesize
\begin{tabular}{lcccccccc}
\toprule
Model & Rollout & Physics & MAE & RMSE & MAPE\% & S1 MAE & S2 MAE & S3 MAE \\
\midrule
PIWM            & Yes & Yes & \textbf{0.0063} & 0.0097 & 0.66 & 0.0056 & \textbf{0.0080} & 0.0185 \\
WM              & Yes & No  & \textbf{0.0063} & \textbf{0.0090} & \textbf{0.65} & 0.0054 & 0.0098          & \textbf{0.0135} \\
CNN-PatchTST    & No  & No  & 0.0078          & 0.0112 & 0.81 & 0.0071 & 0.0103          & 0.0149 \\
\bottomrule
\end{tabular}}
\end{table}

Two findings emerge from the architecture ablation.

Removing the dynamics rollout increases overall MAE from 0.0063 to
0.0078, a 24\% degradation, and doubles the forecast error at short
horizons as shown in Table~\ref{tab:future}. The rollout provides a
multitask learning benefit even for current SOH estimation. Training
on the trajectory objective forces the encoder to produce a richer
latent representation than what current step estimation alone
requires.

PIWM and WM achieve identical aggregate MAE of 0.0063. RMSE separates
them: WM achieves 0.0090 versus 0.0097 for PIWM, indicating that the
physics constraint introduces larger outlier errors. Per-stage results
locate the effect. PIWM achieves lower error at Stage~2, the
degradation knee, with MAE 0.0080 compared to 0.0098 for WM. WM is
better at Stage~3, late degradation, with MAE 0.0135 compared to
0.0185 for PIWM. The physics constraint acts as a regulariser at the knee region,
reducing Stage~2 error by enforcing monotonic degradation where
capacity fade accelerates.

\subsection{Future SOH Trajectory Forecast}

Table~\ref{tab:future} reports future trajectory MAE at horizons 5,
10, 20, and 50 cycles ahead for the architecture variants.

\begin{table}[htbp]
\centering
\caption{Future SOH trajectory MAE at forecast horizons $h \in \{5,
  10, 20, 50\}$ cycles ahead.}
\label{tab:future}
\footnotesize
\begin{tabular}{lcccc}
\toprule
Model & h5 & h10 & h20 & h50 \\
\midrule
PIWM            & 0.0067          & 0.0072          & 0.0081          & 0.0109 \\
WM              & \textbf{0.0065} & \textbf{0.0068} & \textbf{0.0075} & \textbf{0.0096} \\
CNN-PatchTST    & 0.0136          & 0.0135          & 0.0133          & 0.0134 \\
\bottomrule
\end{tabular}
\end{table}

The critical comparison is between the rollout models and
CNN-PatchTST. PIWM achieves MAE of 0.0067 at horizon~5 versus
0.0136 for CNN-PatchTST, a factor of two improvement. This gap is the
core evidence that latent space dynamics rollout produces meaningfully
better forecasts at short horizons than direct regression from a
static latent vector.

CNN-PatchTST shows flat forecast error across all horizons, ranging
from 0.0133 to 0.0136. This is consistent with a model that has no
mechanism to propagate degradation forward and instead learns a
single average trajectory slope applied uniformly. The rollout models
show error growing with horizon, from 0.0067 at horizon~5 to 0.0109
at horizon~50, consistent with uncertainty compounding over the
rollout window. This growing error profile is a signature of genuine
iterative forecasting.

WM achieves slightly better future horizon MAE than PIWM at all
horizons, with horizon~50 MAE of 0.0096 versus 0.0109. This is
consistent with the physics constraint improving current SOH
estimation at the knee region without benefiting trajectory
forecasting.

\subsection{Trajectory Visualisation}

\begin{figure}[htbp]
  \centering
  \safeinclude[width=\textwidth]{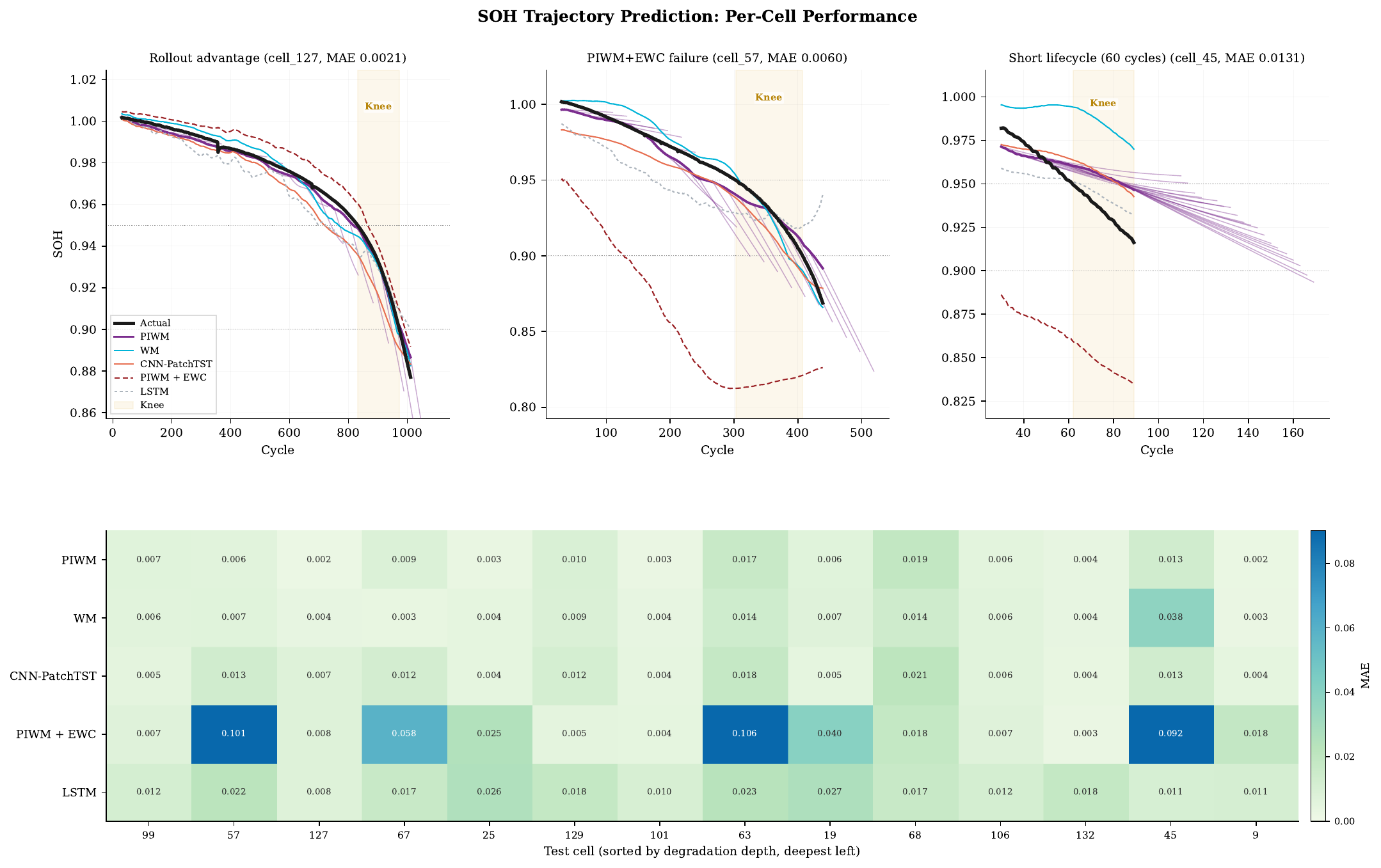}{SOH trajectory prediction: hero panels and heatmap}
  \caption{Per-cell SOH trajectory performance. Top row: three
    representative cells showing rollout advantage on cell~127,
    PIWM+EWC failure mode on cell~57, and short lifecycle limitation
    on cell~45. Forecast wicks in purple show 80 cycles ahead
    predictions from PIWM, 20 wicks per cell. Bottom:
    MAE heatmap for all five methods across all 14 test cells, sorted
    by degradation depth.}
  \label{fig:trajectory}
\end{figure}

Figure~\ref{fig:trajectory} presents per-cell trajectory performance
in two complementary views. The top row shows three representative
cells selected to illustrate distinct model behaviours. Cell~127 shows the largest rollout advantage, with PIWM
achieving MAE 0.002 compared to 0.008 for CNN-PatchTST. Cell~57
exposes the PIWM+EWC failure mode, where batch-staged training
produces MAE 0.101 compared to 0.006 for PIWM with joint training.
Cell~45 is the shortest lifecycle cell at 169~cycles, where the
input window of 30 cycles and verification horizon of 80 cycles
leave only 59 evaluable points. Cell~45 reaches SOH~0.90 at
cycle~100, faster than any cell in the training set where the
earliest SOH~0.90 crossing occurs at cycle~214. The model
extrapolates beyond its training distribution on this cell.
Forecast wicks, 20 per cell and evenly spaced, show predictions
from PIWM 80 cycles ahead, anchored to the prediction line at each
origin cycle.

The heatmap in Figure~\ref{fig:trajectory} reports MAE for all five
methods across all 14 test cells, sorted by degradation depth with
the most degraded cells on the left. This exposes failure modes at
the cell level that are invisible in aggregate metrics. PIWM
achieves MAE below 0.005 on 10 of 14 cells. The highest errors
occur on cell~45, the only test cell that degrades outside the
training distribution, and on cells where PIWM+EWC fails
catastrophically due to batch-staged training.

Every cell loses 110 cycles from evaluation. The first 30 cycles are
consumed as the input window, and the last 80 cycles lack sufficient
ground truth to verify the full forecast horizon.

\subsection{Control Tests}

\subsubsection{Continual Learning}

Table~\ref{tab:ewc_batches} details the batch-staged EWC training
progression, showing how performance evolves as each manufacturing
batch is incorporated. The same PIWM architecture is used and only
the training strategy differs.

\begin{table}[htbp]
\centering
\caption{PIWM + EWC batch-staged training progression. Validation
  MAE at the end of each batch phase.}
\label{tab:ewc_batches}
\footnotesize
\begin{tabular}{lcc}
\toprule
Phase & Validation MAE & Early stop epoch \\
\midrule
Batch~1 & 0.023 & 18 \\
Batch~2 & 0.057 & 18 \\
Batch~3 & 0.014 & 20 \\
\midrule
Test & 0.021 & --- \\
\bottomrule
\end{tabular}
\end{table}

The final test MAE of 0.021 is 3.3 times worse than PIWM with
joint training at 0.006. The three manufacturing batches in the
Severson dataset share the same cell chemistry, format, and
degradation task. When sequential batches share the same data
distribution, the catastrophic forgetting problem that EWC addresses
does not arise. Joint training on the combined dataset is strictly
superior.

The forgetting analysis shows partial protection. Batch~1 test MAE
degrades from 0.004 immediately after batch~1 training to 0.005
after batch~2 and 0.013 after batch~3. EWC slows but does not
prevent knowledge erosion across batches.

\subsubsection{LSTM Baseline}

The LSTM baseline achieves overall MAE~0.0209, 3.3 times worse
than PIWM at 0.0063. Table~\ref{tab:controls} summarises both
control tests alongside PIWM for reference.

\begin{table}[htbp]
\centering
\caption{Control tests compared with PIWM under joint training.}
\label{tab:controls}
\resizebox{\columnwidth}{!}{%
\footnotesize
\begin{tabular}{lccccccc}
\toprule
Model & Training & MAE & RMSE & MAPE\% & S1 MAE & S2 MAE & S3 MAE \\
\midrule
PIWM             & Joint        & \textbf{0.0063} & \textbf{0.0097} & \textbf{0.66} & 0.0056 & 0.0080 & 0.0185 \\
PIWM + EWC       & Batch-staged & 0.0210          & 0.0377          & 2.19          & 0.0167 & 0.0413 & 0.0341 \\
LSTM             & Joint        & 0.0209          & 0.0286          & 2.20          & 0.0164 & 0.0344 & 0.0598 \\
\bottomrule
\end{tabular}}
\end{table}

\subsection{Latent Space Analysis}

\begin{figure}[htbp]
  \centering
  \begin{subfigure}[t]{0.48\textwidth}
    \safeinclude[width=\textwidth]{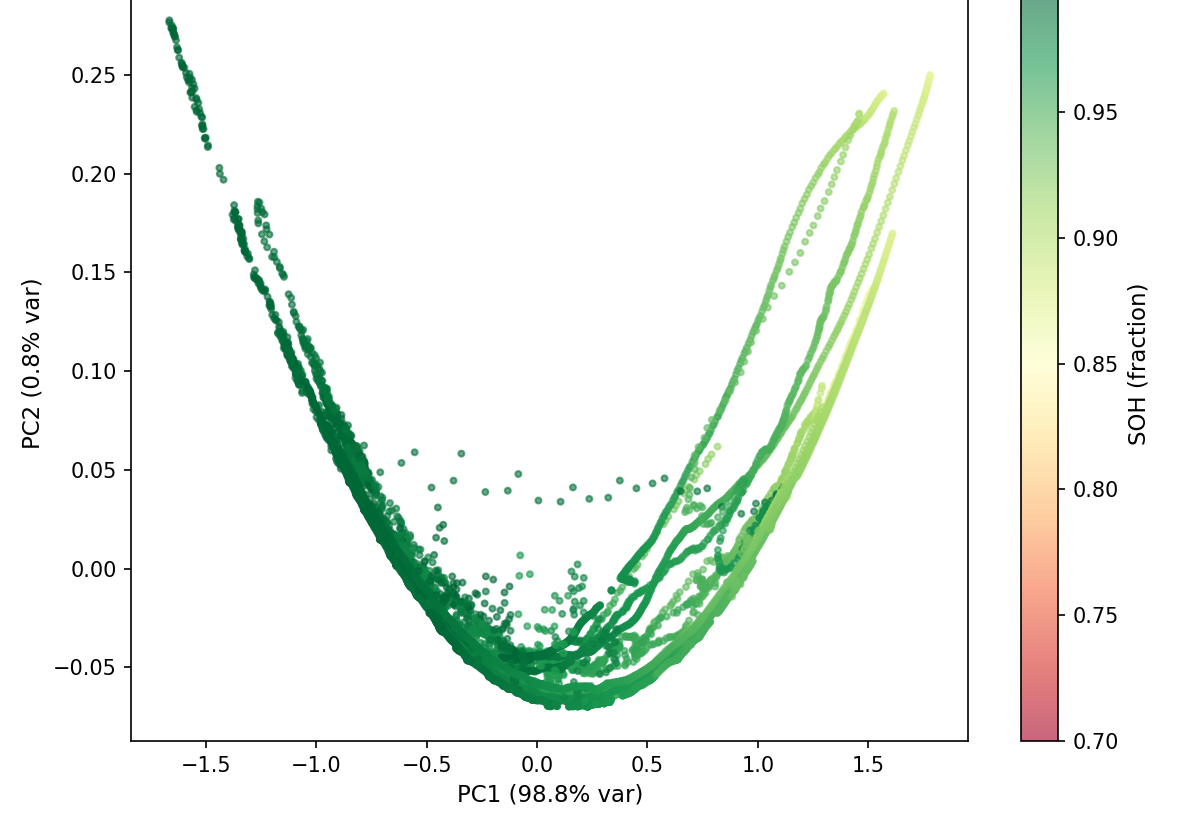}{CNN-PatchTST latent PCA}
    \caption{CNN-PatchTST: per-cell strands, no shared trajectory.}
  \end{subfigure}
  \hfill
  \begin{subfigure}[t]{0.48\textwidth}
    \safeinclude[width=\textwidth]{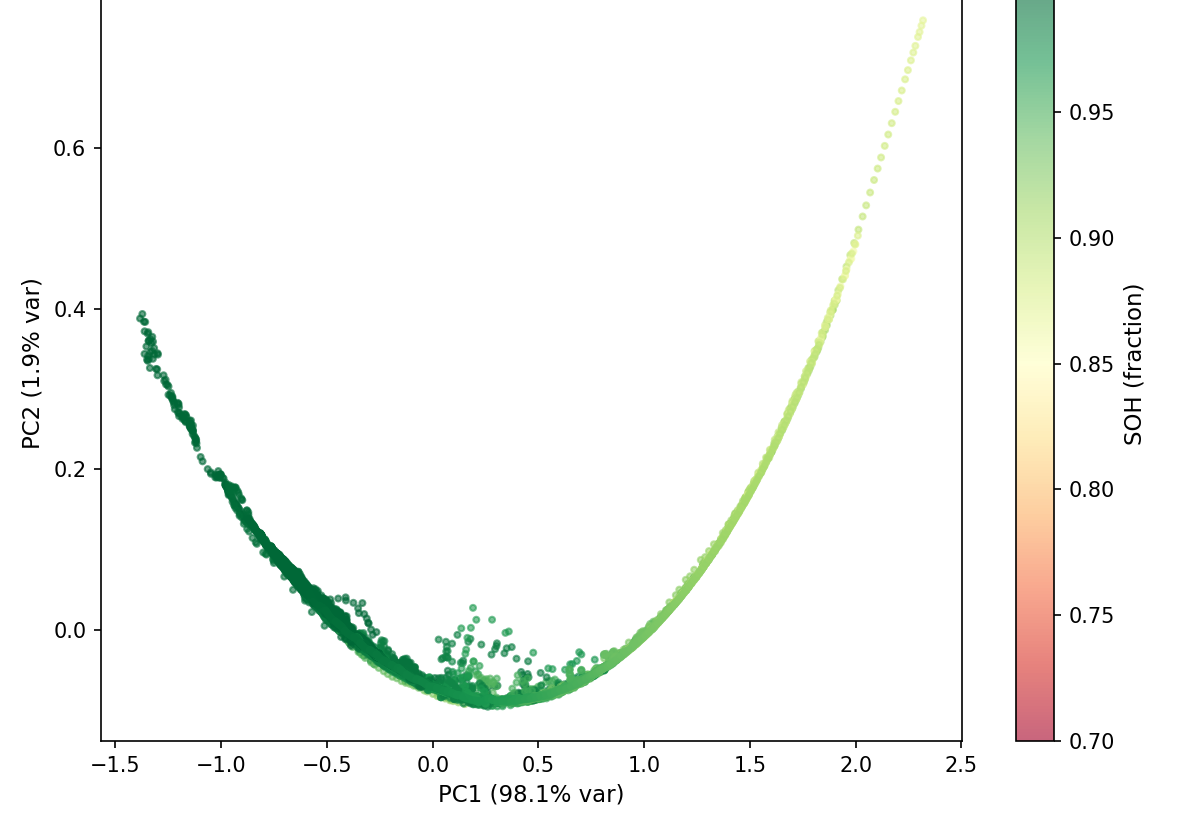}{WM latent PCA}
    \caption{WM: unified curve, SOH ordered (PC2 1.9\%).}
  \end{subfigure}
  \\[6pt]
  \begin{subfigure}[t]{0.48\textwidth}
    \safeinclude[width=\textwidth]{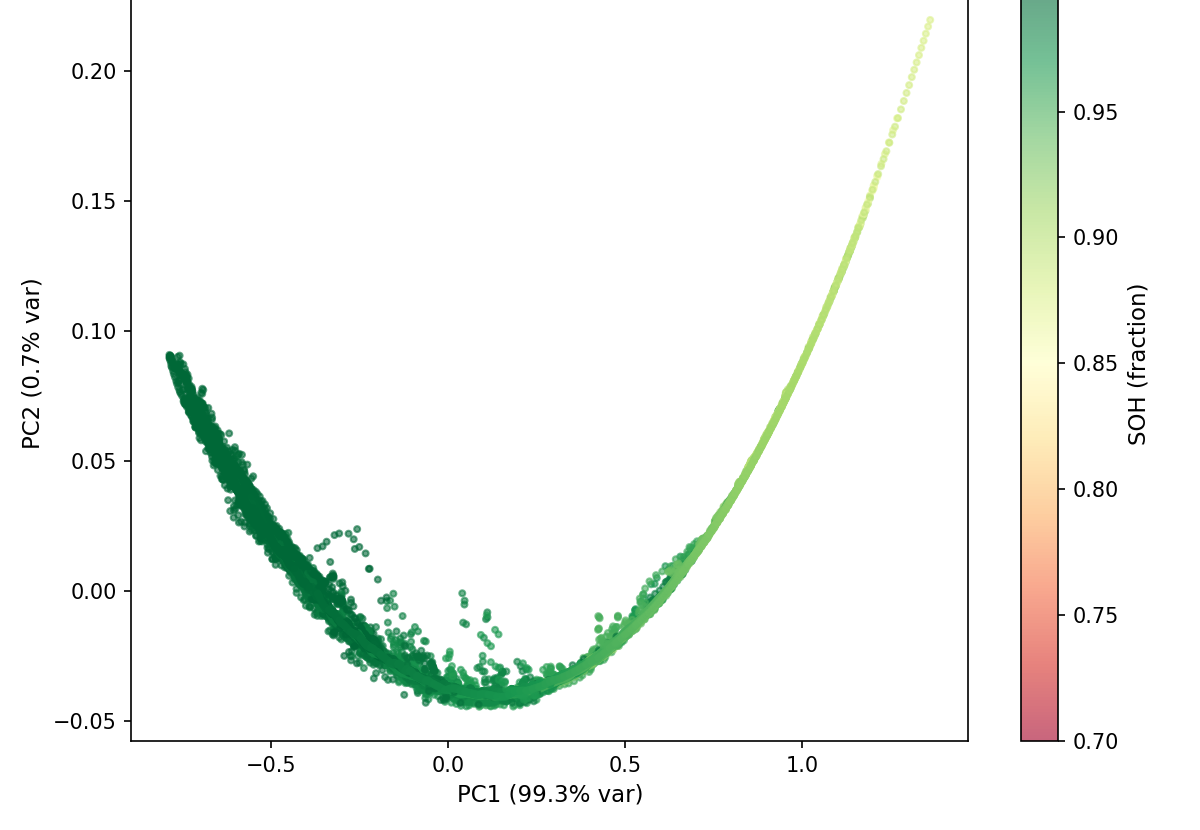}{PIWM latent PCA}
    \caption{PIWM: unified curve, similar to WM (PC2 0.7\%).}
  \end{subfigure}
  \hfill
  \begin{subfigure}[t]{0.48\textwidth}
    \safeinclude[width=\textwidth]{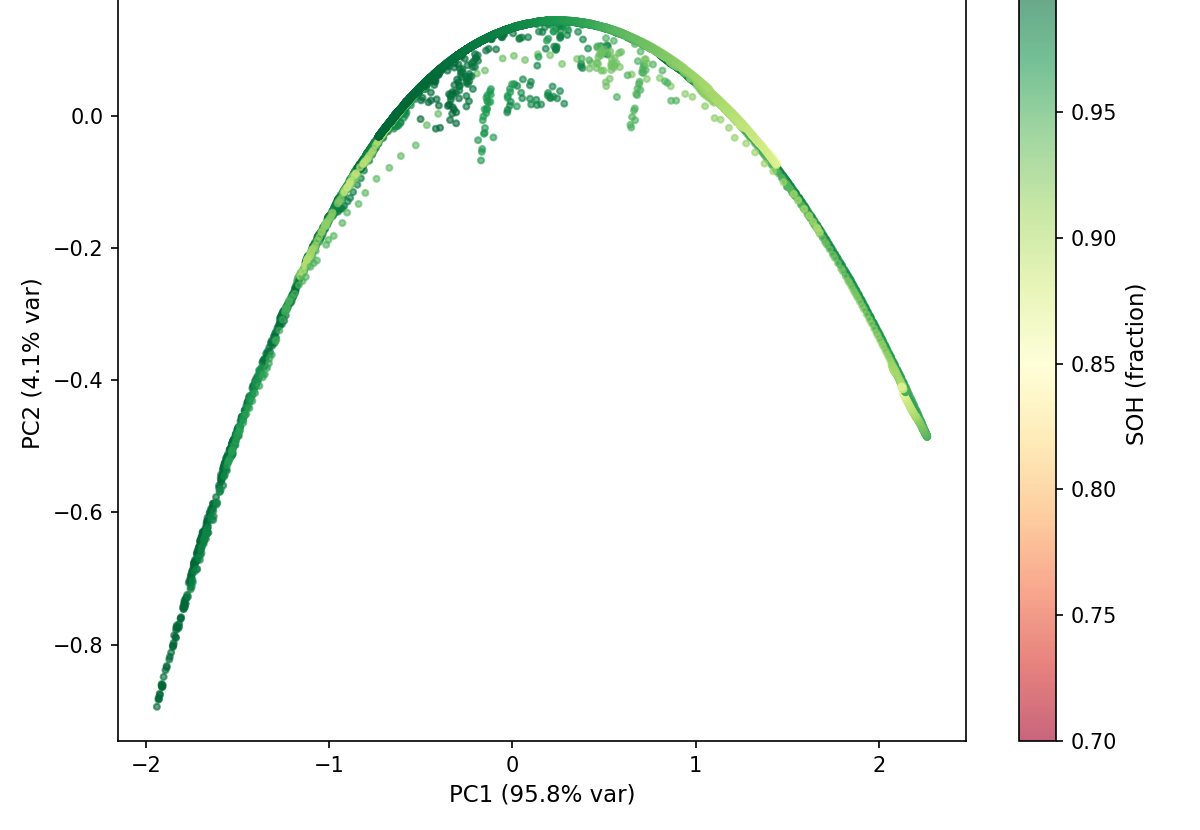}{PIWM+EWC latent PCA}
    \caption{PIWM + EWC: distorted manifold from batch-staged training.}
  \end{subfigure}
  \caption{PCA of latent vectors $\zk$, coloured by SOH value.
    The rollout objective produces a unified SOH-ordered curve (b, c),
    while the encoder without rollout retains per-cell structure (a)
    and batch-staged training distorts the manifold (d).}
  \label{fig:pca}
\end{figure}

Figure~\ref{fig:pca} shows PCA projections of the latent vectors
$\zk$ for the four variants that share the CNN-PatchTST encoder,
coloured by SOH value. CNN-PatchTST shows
multiple separated strands corresponding to individual cells,
indicating the encoder retains cell identity rather than learning a
shared degradation representation. Both rollout models, PIWM and WM,
compress all cells onto a single tight curve ordered by SOH. The
transition function requires a unified trajectory to step along, and
the rollout training objective forces the encoder to produce this
structure. PIWM and WM show similar latent geometry, indicating that
the physics constraint does not reshape the latent space. PIWM + EWC
shows an inverted arch with more variance in PC2 (4.1\% versus
0.7--1.9\% for the other variants), indicating that batch-staged
training disrupts the latent organisation that joint training
produces.

\section{Discussion}
\label{sec:discussion}

\subsection{Effect of the Dynamics Rollout}

The rollout improves both current estimation and future forecasting
through a multitask learning effect. The rollout objective forces the
encoder to produce latent states that encode not only current SOH but
also how SOH will evolve. The shared decoder then benefits from this
enriched representation even for the current step estimate.

The latent space PCA in Figure~\ref{fig:pca} provides direct evidence.
CNN-PatchTST, trained without rollout, produces per-cell strands in
latent space where the encoder retains cell identity. The rollout
models compress all cells onto a single SOH-ordered curve. The
transition function requires a unified trajectory to step along, and
the training objective enforces this structure. This explains why
CNN-PatchTST produces flat forecast error across horizons: without a
shared trajectory in latent space, the decoder learns a single
average slope rather than iterative propagation.

\subsection{Physics Constraint as Targeted Regularisation}

The monotonicity penalty is most effective at the knee where
capacity fade accelerates and the model is prone to predicting
temporary SOH recovery. Enforcing nonincreasing trajectories
reduces Stage~2 MAE. Below SOH 0.90, the constraint
overregularises at Stage~3, increasing both MAE and RMSE.

The RMSE difference between PIWM and WM (0.0097 versus 0.0090)
indicates that the physics penalty introduces larger outlier errors.
These outliers concentrate at Stage~3 where the constraint is least
valid. WM also achieves better future trajectory MAE at all horizons,
consistent with the physics penalty adding a small bias that
compounds over the rollout window.

The physics constraint weight $\lambda_{\mathrm{phys}} = 0.1$ is
untuned. Stage-dependent activation or adaptive balancing such as
GradNorm~\cite{gradnorm2018} may improve the tradeoff between
Stage~2 and Stage~3. This is deferred to subsequent work.

\subsection{Continual Learning and Data Distribution}

The three manufacturing batches in the Severson dataset share
the same cell chemistry, format, and degradation task. When
sequential batches are drawn from the same distribution, the
catastrophic forgetting problem that EWC addresses does not arise.
Joint training is strictly superior because the batches contain
redundant, not conflicting, information.

Continual learning would be better motivated when sequential data
represents genuinely different distributions, such as different cell
chemistries or different operating conditions. In such settings, new
data can destructively interfere with previously learned
representations, and EWC parameter protection becomes necessary.

The PCA in Figure~\ref{fig:pca}(d) shows that batch-staged training
distorts the latent manifold, with PC2 capturing 4.1\% of variance
compared to 0.7--1.9\% under joint training. The sequential
optimisation path produces a qualitatively different latent geometry
that the transition function cannot compensate for.

\subsection{Generalisation Boundary}

Cell~45 degrades faster than any cell in the training set, reaching
SOH~0.90 at cycle~100 versus cycle~214 for the fastest training
cell. All methods produce higher error on this cell because they
extrapolate beyond the training distribution. This is not a model
failure but a data coverage boundary: the model cannot be expected to
generalise to degradation rates it has never observed.

The test set contains 14 cells with lifecycles ranging from 169 to
1835 cycles. The stratified split by degradation depth ensures
coverage across degradation stages, but does not guarantee coverage
of all degradation rates. Cell~45 exposes this gap.

\subsection{Limitations}

The contribution of raw time-series input via the cycle encoder
versus summary statistics is not isolated in the current ablation.

The test set contains 14 cells. Stage~3 results are computed over
fewer samples and should be interpreted with caution.

All experiments use the Severson LFP dataset. Generalisability to
other cell chemistries such as NMC and LCO and to other cycling
protocols is not tested. The A123 LFP cell exhibits a relatively
flat voltage plateau, which may make the discharge curve shape less
informative than for chemistries with more pronounced voltage
features such as NMC and NCA.

\section{Conclusion}
\label{sec:conclusion}

This paper formulated battery degradation prognosis as a world model
problem and studied the contribution of iterative latent rollout, a
physics constraint, and continual learning on the
Severson LFP dataset of 138~cells. The dynamics rollout is the
essential component: it halves the forecast error at short horizons
and forces the encoder to learn a unified latent trajectory that
benefits both SOH estimation and trajectory forecasting. The physics constraint improves
prediction at the degradation knee at the cost of late degradation
accuracy. EWC continual
learning provides no benefit when sequential batches share the same
data distribution. Future work includes adaptive physics constraint
balancing, extension to additional cell chemistries, and continual
learning across heterogeneous battery populations.

\bibliographystyle{unsrtnat}
\bibliography{references}

\end{document}